%% file: main.tex
\documentclass[letterpaper]{article} 
\usepackage{aaai24}  
\usepackage{times}  
\usepackage{helvet}  
\usepackage{courier}  
\usepackage[hyphens]{url}  
\usepackage{graphicx} 
\def\UrlFont{\rm}  
\usepackage{natbib}  
\usepackage{caption} 
\usepackage{algorithm}
\usepackage{algorithmic}

\usepackage{newfloat}
\usepackage{listings}
\DeclareCaptionStyle{ruled}{labelfont=normalfont,labelsep=colon,strut=off} 
\floatstyle{ruled}
\newfloat{listing}{tb}{lst}{}
\floatname{listing}{Listing}
\usepackage{url}
\usepackage{placeins}
\usepackage{tikz}
\usetikzlibrary{arrows.meta, positioning, quotes, fit,shapes.geometric}
\usepackage{xcolor}
\usepackage{color}
\usepackage{float}
\usepackage{enumitem}
\usepackage{pgfplots}
\usepackage{graphicx}
\usepackage{url}
\definecolor{navy}{RGB}{0,0,128}

\definecolor{darkgreen}{rgb}{0.0, 0.5, 0.0}
\definecolor{amaranth}{rgb}{0.9, 0.17, 0.31}
\definecolor{azure}{rgb}{0.0, 0.5, 1.0}
\newif\ifcommenton
 \commentontrue

\ifcommenton
\newcommand{\TODO}[1]{\textcolor{red}{[TODO] #1}}
\newcommand{\hugo}[1]{{\color{brown}\bfseries [Hugo: #1]}}
\newcommand{\shan}[1]{{\color{orange}\bfseries [Shan: #1]}}
\newcommand{\payman}[1]{{\color{violet}\bfseries [Payman: #1]}}
\newcommand{\alexey}[1]{\textcolor{darkgreen}{[AT: #1]}}
\newcommand{\alind}[1]{\textcolor{darkgreen}{[Alind: #1]}}
\newcommand{\mike}[1]{{\color{purple}\bfseries [Mike: #1]}}
\newcommand{\hanning}[1]{{\color{olive}\bfseries [Hanning: #1]}}
\newcommand{\fixme}[1]{{{\color{blue} #1}}}
\else
\newcommand{\TODO}[1]{}
\newcommand{\hugo}[1]{}
\newcommand{\shan}[1]{}
\newcommand{\payman}[1]{}
\newcommand{\alexey}[1]{}
\newcommand{\alind}[1]{}
\newcommand{\hanning}[1]{}
\newcommand{\mike}[1]{}
\newcommand{\fixme}[1]{}
\fi

\begin{document}
\input{sections/titleandauthors.tex}

\maketitle
\input{sections/abstract.tex}
\input{sections/introduction.tex}
\input{sections/relatedWork.tex}

\input{sections/algorithmOverview}
\input{sections/ExperimentsAndResults}
\input{sections/conclusionAndNextSteps}
\bibliography{bibliography/Library_Ethosight, bibliography/Library_Hugo, bibliography/Library_Kris, bibliography/addtionalRefs, bibliography/aaai24}
\end{document}

%% file: sections/titleandauthors.tex


\include{sections/title}

 \author{
     Hugo Latapie\textsuperscript{\rm 1}, Shan Yu\textsuperscript{\rm 1}, Patrick Hammer\textsuperscript{\rm 3}, Kristinn R. Th{\'o}risson\textsuperscript{\rm 2}, Vahagn Petrosyan\textsuperscript{\rm 1}, Brandon Kynoch\textsuperscript{\rm 1}, Alind Khare\textsuperscript{\rm 6}, Payman Behnam\textsuperscript{\rm 6}, Alexey Tumanov\textsuperscript{\rm 6},Aksheit Saxena\textsuperscript{\rm 5},Anish Aralikatti\textsuperscript{\rm 5}, Hanning Chen\textsuperscript{\rm 7}, Mohsen Imani\textsuperscript{\rm 7} , Mike Archbold\textsuperscript{\rm 8}, Tangrui Li\textsuperscript{\rm 4},Pei Wang\textsuperscript{\rm 4}, Justin Hart\textsuperscript{\rm 9}
 }

 \affiliations{ 
     \textsuperscript{\rm 1} Cisco Research, San Jose, CA \\
     \textsuperscript{\rm 3} Department of Psychology, Stockholm University, Sweden \\
     \textsuperscript{\rm 2} Icelandic Institute for Intelligent Machines and CADIA, Reykjavik University, Iceland  \\
     \textsuperscript{\rm 6} Georgia Institute of Technology \\
      \textsuperscript{\rm 5} Cisco Systems, Bengaluru, India\\
      \textsuperscript{\rm 7} University of California, Irvine \\
      \textsuperscript{\rm 8} AbsoluteMode Engine \\
     \textsuperscript{\rm 4} Department of Computer and Information Sciences, Temple University, USA \\
     \textsuperscript{\rm 9} The University of Texas at Austin \\

 }

%% file: sections/title.tex


\title{Ethosight:~Reasoning-Guided Hybrid AI for Nuanced Perception \\ via Joint-Embedding \& Multi-modal Contextual Affinity}

%% file: sections/abstract.tex
\begin{abstract}
Traditional computer vision models often necessitate extensive data acquisition, annotation, and validation. These models frequently struggle in real-world applications, resulting in high false positive and negative rates, and exhibit poor adaptability to new scenarios, often requiring costly retraining. To address these issues, we present Ethosight, a flexible and adaptable zero-shot video analytics system. Ethosight begins from a clean slate based on user-defined video analytics, specified through natural language or keywords, and leverages joint embedding models and reasoning mechanisms informed by ontologies such as WordNet and ConceptNet. Ethosight operates effectively on low-cost edge devices and supports enhanced runtime adaptation, thereby offering a new approach to continuous learning without catastrophic forgetting. We provide empirical validation of Ethosight's promising effectiveness across diverse and complex use cases, while highlighting areas for further improvement. A significant contribution of this work is the release of all source code and datasets to enable full reproducibility and to foster further innovation in both the research and commercial domains.
\end{abstract}

%% file: sections/introduction.tex
\section{Introduction}
Computer vision technologies are rapidly advancing and increasingly employed in real-world applications for safety, security, and health analytics. However, conventional models often struggle to detect subtle, context-dependent behaviors and events, due to the high variability of real-world scenarios. These challenges typically necessitate extensive data acquisition, laborious annotation, and validation processes. Moreover, traditional approaches frequently demand significant computational resources and time for training, thereby increasing the cost overhead substantially.

Furthermore, despite these investments, these models often do not generalize well when exposed to new, varying scenarios. This poor generalization can lead to high false positive rates, where benign activities are incorrectly flagged as incidents, and high false negative rates, where actual critical events are missed. This lack of adaptability compromises the reliability of the system and significantly limits its utility and cost-effectiveness, putting organizations in a perpetual loop of costly refinement.

To address these issues, we present Ethosight, a video analytics system designed to minimize the traditional barriers to effective and efficient computer vision. Unlike traditional models, Ethosight initiates from a clean slate based on user-defined video analytics, specified in natural language or through specific keywords of interest.

Ethosight operates in multiple modes, tailored to the diverse needs of real-world deployments. In one mode, Ethosight compiles user-defined video analytics into an optimized package. This package is designed to be executed efficiently on low-cost edge devices, empowering real-time, data-driven insights at the edge. To create this package, Ethosight leverages reasoning mechanisms and prior knowledge, sourced from ontologies such as WordNet and ConceptNet\cite{Latapie2022}\cite{latapie2021}.

Ethosight also supports enhanced runtime adaptation. It follows a reasoning process similar to the initial phase, producing analytics with associated confidence scores. These scores can trigger further refinement of the model during runtime as needed, allowing Ethosight to adapt to evolving conditions without extensive retraining.

Unique to Ethosight is its generation of symbolic knowledge in graph form. This feature enables cumulative and continuous learning while avoiding the catastrophic forgetting that plagues traditional machine learning systems. By storing new learnings in this symbolic form, Ethosight facilitates knowledge sharing among various deployments, enhancing the value and adaptability of each system over time.

At the heart of Ethosight's operation is its use of pre-trained models. Among various options, we have found ImageBind, a multi-modal joint embedding model, to be particularly effective for generating precise and nuanced semantic knowledge of images. This has proven to be invaluable for the initial preparation of Ethosight's “compiled” analytics packages as well as optional run-time learning and adaptation capability. However, Ethosight is designed to be versatile; it is also compatible with other pre-trained models, such as OpenClip, offering users flexibility based on their specific needs and resources.

For runtime inference, especially in resource-constrained environments like edge devices, we have observed that joint embeddings trained on fewer modalities perform remarkably well. They strike a compelling balance between efficiency and accuracy, thereby enabling Ethosight to deliver actionable insights in real-time, even when computational resources are limited.

A distinguishing feature of Ethosight is its incorporation of Korzybski's "time-binding" concept \cite{korzybski1921}\cite{korzybski1994}. This philosophical framework is applied in a practical manner, enabling generational learning and knowledge sharing across different deployments of the system. This not only allows Ethosight to continuously evolve and adapt but also facilitates a form of ‘collective intelligence’ among various Ethosight deployments, akin to human knowledge accumulation and transfer.

Our evaluations focused on a range of use cases utilizing an open dataset comprising images of complex events, which we are introducing in this work. Additionally, we conducted extensive validation using the UCF Crime dataset. Our results are promising; they suggest that zero-shot complex video analytics capabilities are within reach using our approach. However, it is important to note that the performance of Ethosight, while encouraging, does not yet match the state-of-the-art in all aspects. These tests, conducted in diverse environments, serve as valuable empirical evidence for Ethosight's potential and establish a foundation upon which we intend to continually improve and refine the system.

Contributions of this paper include: (1) The introduction of Ethosight, a flexible and adaptable zero-shot video analytics system; (2) A novel approach to generating symbolic knowledge, enabling continuous learning without catastrophic forgetting; (3) Empirical validation of Ethosight’s effectiveness across diverse and complex use cases, with results that are promising while acknowledging areas for further improvement; (4) The release of all source code and datasets associated with this work, to promote full reproducibility of our results and encourage further innovation in both the research and commercial domains.

%% file: sections/relatedWork.tex
\section{Related Work}
\subsection{Image Classification}
Substantial works have investigated the enhancement of accuracy in image/video classification. Vasudevan et al. discovered that DNN models tend to have errors on images that are ambiguous or contain multiple objects \cite{vasudevan2022does}. Yun et al. noted that many images in the original ImageNet dataset contain multiple objects while the dataset is single-labeled. Such images with multiple objects could present challenges, as models might struggle to concentrate on the pertinent objects. As a result, they advocated for re-labeling the ImageNet training set with multi-labels~\cite{yun2021re}.
In another study, Luccioni and Rolnick pointed out that  many of the classes especially animal ones are ill-defined or overlapping~\cite{luccioni2023bugs}. This skewed distribution in biodiversity can introduce biases, leading to inaccurate predictions by machine learning models. Another study suggested utilizing category-specific visual-semantic mappings and label refinement to enhance the accuracy of ImageNet prediction~\cite{niu2018zero}. Nevertheless, none of these studies have the ability of nuance prediction. 
\subsection{Zero-shot Learning}
Zero-shot learning is a transformative paradigm in machine learning, where the goal is to recognize objects without having previously seen examples of them during training. Significant contributions to zero-shot learning have been made in the past and can achieve state-of-the-art accuracy on diverse benchmarks. 
 Goyal et al. \cite{Goyal_2023_CVPR} highlight that in zero-shot learning even subtle changes in the fine-tuning process can have surprisingly large impacts on both in-distribution and out-of-distribution data performance. They propose a contrastive fine-tuning method, known as FLYP, which continues to optimize the contrastive loss between image embeddings and class-descriptive prompt embeddings. This approach consistently outperforms other fine-tuning methods across multiple benchmarks, including distribution shift, transfer learning, and few-shot learning scenarios. This is consistent with the observations made by Kumar et al. on the sensitivity of these models to the fine-tuning process~\cite{kumar2022calibrated}. 


Parallel lines of research have investigated the applicability of zero-shot learning beyond the field of computer vision, focusing on the importance of deciphering and transferring the mapping relationships between the features and labels of seen classes to unseen classes~\cite{cao2020research}. This exploration of zero-shot learning in diverse fields augments our understanding of its potential applications. 

Central to many zero-shot learning approaches are large embeddings. Our proposed approach, Ethosight, adopts ImageBind as its backbone multi-modal embedding model~\cite{girdhar2023imagebind}.
ImageBind works by learning a joint embedding space for six different modalities: images, text, audio, depth, thermal, and IMU data. The neural network learns to map each data point from each modality to a single vector in the joint embedding space. While Ethosight builds upon ImageBind, it makes unique contributions by incorporating a self-supervised exploration strategy and localized label affinity calculations, thereby enabling the extraction of subtle and nuanced relationships within given contexts.

Nevertheless, these approaches differ from our proposed Ethosight model, which employs localized label affinity calculations for label discovery and iterative refinement. Uniquely, Ethosight's method facilitates the extraction of subtle and nuanced relationships within a given context, an aspect that extends its applicability beyond traditional paradigms and into more complex and nuanced scenarios. This distinguishing approach contributes to enhancing the broad-spectrum utility of zero-shot learning strategies.

\subsection*{Visual-based Video Anomaly Detection}

The literature on employing visual features for video anomaly detection can be broadly segmented into two paradigms based on the nature of the training data: unsupervised learning and weakly-supervised learning approaches.

\textbf{Unsupervised Learning Approaches:} These methods are based on the fact that only normal videos are available during training and transform the anomaly detection task into a reconstruction problem. Unsupervised learning approaches are based on the assumption that the models are only trained to learn the representation of normal videos, making it hard for abnormal videos to reconstruct.
Classic features such as the histogram of oriented gradients \cite{wang2009hog} and histogram of optical flow \cite{dalal2006human}, are commonly employed. Additionally, visual features extracted using deep learning mechanisms like CNN \cite{rezaei2021new} have been advocated. To increase performance, various reconstructive tools, including deep autoencoders \cite{wang2020abnormal}, and GANs \cite{wang2021video} have been employed. Despite their advancements, unsupervised approaches still face challenges in delivering satisfactory abnormal detection outcomes on real-world datasets \cite{sultani2018real}, primarily due to the absence of prior knowledge of abnormal data.

\textbf{Weakly-supervised Learning Approaches:} These methods necessitate the use of video-level labels indicating "normal" or "abnormal" for training. Sultani et al. \cite{sultani2018real} proposed the multiple instance learning (MIL) approach, which achieves high abnormal detection performance by maximizing the discrepancy between the anomaly scores of positively and negatively labeled video segments. Tian et al. \cite{tian2021weakly} enhanced MIL through their robust temporal feature magnitude (RTFM) learning technique, selecting the top-k segments from a video rather than relying on a singular segment. In addition, various MIL-driven methodologies \cite{wu2021learning, panariello2022consistency,lv2021localizing} incorporate diverse functions for frame aggregation and feature extraction. Nonetheless, these weakly-supervised methods mandate the accumulation and annotation of datasets encompassing abnormal scenarios, which demands significant human effort.
\vspace{-2 ex}
\subsection{Multi-modal-based Video Anomaly Detection}
In addition to visual features, a few recent studies leverage text features for semantic information, so as to enhance video anomaly detection.
For instance, Chen et al. \cite{chen2023tevad} introduced the TEVAD framework, which computes text features from video captions, and extends multi-scale temporal learning to text features to better capture the dependencies between snippet features.
However, the TEVAD model still relies on weakly-supervised video-level labels. In contrast,  Simmons and Vasa ~\cite{simmons2023garbage} implement a zero-shot video anomaly detection and classification mechanism by tapping into the reasoning capacities of large language models. However, the method relies on high-quality video descriptions, where existing automated video-to-text approaches fail to achieve, leading to very low classification performance. Kim et al. \cite{kim2023unsupervised} suggested an unsupervised approach using extensive language models to derive text descriptions and subsequently discern abnormal frames based on the cosine similarity with the produced text descriptions. While this method eradicates the need for labor-intensive manual labeling, it demands additional fine-tuning of the embedding for specific video dataset adaptation and manual oversight to excise poor or irrelevant text descriptions.

%% file: sections/algorithmOverview.tex
\section*{System Architecture and Approach}

\subsection*{Ethosight Compile Phase}

The Ethosight system initiates with the Compile Phase. This phase begins with the input of a natural language description of the desired video analytics functionality and/or a set of keywords. 

\input{diagrams/EthosightArchitectureDiagram}

\begin{itemize}
    \item \textbf{Input Generation:} For example, in a safety-focused scenario, the input might be a detailed description of safety aspects or specific keywords such as ``fire,'' ``intruder,'' or ``fall detection.''
\end{itemize}

\paragraph*{Semantic Expansion:}
Based on these inputs, Ethosight performs a Semantic Expansion using reasoning and prior knowledge. This knowledge can be derived from human-curated sources such as WordNet, or from machine-generated ontologies produced by other Ethosight systems. 

The Semantic Expansion is designed to achieve three key objectives:
\begin{enumerate}
    \item Enhance Positive Evidence Recognition.
    \item Enhance Negative Evidence Recognition.
    \item Discriminate Between Requested Analytics Classes.
\end{enumerate}

\paragraph*{Outcome:} 
The outcome of the Compile Phase is the Ethosight Application, a structured format that holds:
\begin{enumerate}
    \item New Symbolic Information.
    \item Embeddings for All New Labels and Composite Labels.
\end{enumerate}

\subsection*{Ethosight Inference Loop}

The core of the Ethosight algorithm involves the Ethosight Inference Loop. This loop systematically employs a joint embedding model, such as ImageBind or OpenCLIP.

\subsubsection*{Detailed Steps:}
\begin{enumerate}
    \item Seed Label Initialization.
    \item Computing Affinity Scores.
    \item Optional Reasoning Layer.
    \item Iterative Refinement (if Reasoning is Employed).
    \item Convergence Check (if Reasoning is Employed).
\end{enumerate}

\subsection*{Scene Analysis and Convergence (Optional)}
\begin{itemize}
    \item Note: This stage is pertinent when the optional reasoning layer is employed within the Ethosight Inference Loop.
    \item At each iteration of the loop, when reasoning is active, a Scene Analysis is performed.
\end{itemize}

\subsection*{Deep Vision Ethosight Analytics Stream}

\subsubsection*{Ethosight as a Deep Vision Application:}

Ethosight represents the inaugural application developed within the Deep Vision framework, a scalable, real-time stream analytics platform engineered specifically for hybrid AI applications.

\paragraph*{Localized Label Affinities:}
\begin{itemize}
    \item Ethosight can calculate localized label affinities, which involves partitioning the image into a grid (options include 3x3, 4x4, or 5x5) and computing the affinity scores for each grid cell. This step generates a heatmap, offering a more detailed, localized view of the semantic affinities within the image.
\end{itemize}

\paragraph*{Self-Supervised Iterative Learning:}
\begin{itemize}
    \item The iterative learning process, which is invoked when Ethosight operates with its optional reasoning layer, is self-supervised.
\end{itemize}

\subsubsection*{Expanding Deep Vision Applications:}
\begin{itemize}
    \item While Ethosight sets the stage as the first application, the Deep Vision framework is designed to be extensible. Future applications under development include hybrid AI object trackers.
\end{itemize}

%% file: diagrams/EthosightArchitectureDiagram.tex
\begin{figure}[ht]
\centering
\resizebox{0.8\columnwidth}{!}{
\begin{tikzpicture}[
  node distance = 5mm and 2mm,
  block/.style = {draw, rectangle, minimum width=15mm, minimum height=7mm, align=center},
  arrow/.style = {->, >=stealth},
  font=\scriptsize
]

\node[block] (input) {User Input};
\node[block, below=of input, yshift=-5mm] (compile) {Ethosight Compile\\ Phase};

\node[block, right=of compile, xshift=15mm] (appInput) {Ethosight App};
\node[block, below=of appInput] (embedding) {Label Affinity};
\node[block, below=of embedding] (reasoning) {Optional Reasoning};
\node[block, below=of reasoning] (sceneAnalysis) {Scene Analysis};
\node[draw, diamond, aspect=2, below=of sceneAnalysis] (decision) {Converged?};
\node[block, dashed, below=of decision] (optional) {Optional: NLG / API};
\node[block, below=of optional, yshift=-1mm] (analytics) {Deep Vision\\ Ethosight \\ Analytics Stream};

\node[block, above=of appInput] (deepvision) {Deep Vision\\ Video Stream};

\draw [arrow] (input) -- (compile);
\draw [arrow] (compile) -- (appInput);
\draw [arrow] (appInput) -- (embedding);
\draw [arrow] (embedding) -- (reasoning);
\draw [arrow] (reasoning) -- (sceneAnalysis);
\draw [arrow] (sceneAnalysis) -- (decision);
\draw [arrow] (decision) -- node[left] {Yes} (optional);
\draw [arrow] (decision) -- node[above] {No} ++(-1.5,0) |- (reasoning);
\draw [arrow] (deepvision) -- (appInput);
\draw [arrow] (optional) -- (analytics);

\node[draw, fit=(appInput) (embedding) (reasoning) (sceneAnalysis) (decision) (optional), inner xsep=3mm, label=above:{Ethosight Inference}] {};

\end{tikzpicture}
}
\caption{Ethosight Overview}
\label{fig:ethosight_system}
\end{figure}
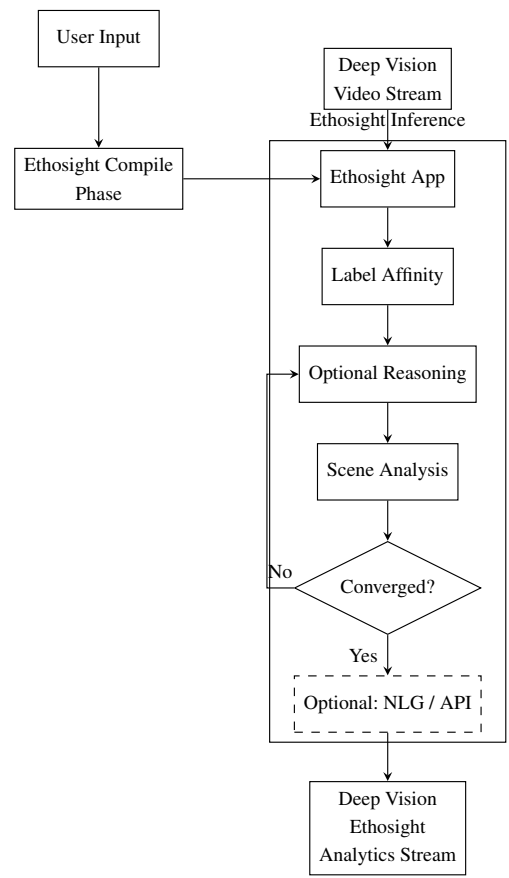

%% file: sections/ExperimentsAndResults.tex
\section{Experiments and Results}
\subsection{Multi-class Classification Accuracy}

\subsubsection{Objective of the Multi-class Classification Test}

The objective of this test was to evaluate how well the Ethosight model can distinguish between 13 different anomaly classes and 1 ``normal event'' class using the UCF Crime dataset.

\subsubsection{Model Used}

Our model is an Ethosight zero-shot application. It starts from a blank slate with no pre-trained knowledge other than the 14 ground truth labels. The underlying joint embedding technique used for this model is ImageBind.

\subsubsection{Preprocessing and Feature Engineering}

\begin{itemize}
    \item No preprocessing steps are applied to the image data. Images are used as-is without normalization, data augmentation, or handling of missing or corrupted data.
    \item The underlying joint embedding used by Ethosight, in this case ImageBind, effectively captures subtle features in the image data.
    \item The label-to-affinity scores generated by Ethosight are critical in the model's sensitivity to these subtle features.
    \item These scores allow Ethosight to dynamically apply post-processing phases that further refine the model's predictions.
\end{itemize}

\subsubsection{Evaluation Metrics and Results}

\begin{table}[ht]
    \centering
    \begin{tabular}{l c}
        \hline
        \textbf{Metric} & \textbf{Score} \\
        \hline
        Accuracy & 92.81\% \\
        Precision & 92.81\% \\
        Recall & 100\% \\
        F1 Score & 96.27\% \\
        Top-1 Accuracy & 25.15\% \\
        Top-5 Accuracy & 52.10\% \\
        Total Predictions & 167 \\
        \hline
    \end{tabular}
    \caption{Ethosight Evaluation Metrics for Multi-class Classification}
    \label{table:eval_metrics}
\end{table}

\begin{figure}[ht]
    \centering
    \includegraphics[width=0.9\columnwidth]{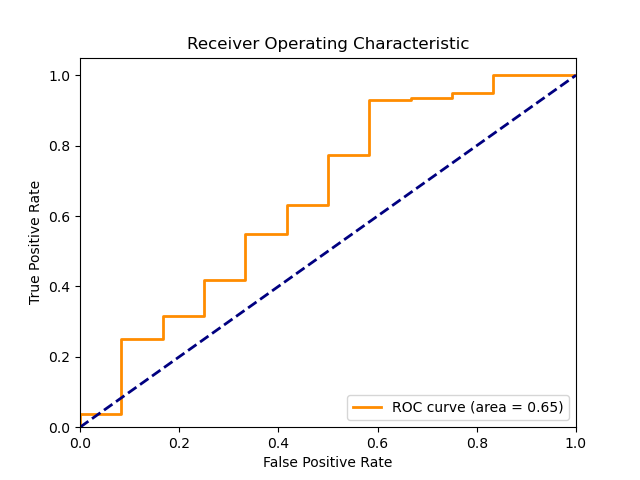}
    \caption{ROC Curve1 - Ethosight Without Semantic Label Expansion}
    \label{fig:roc}
\end{figure}

\begin{figure}[ht]
    \centering
    \includegraphics[width=0.9\columnwidth]{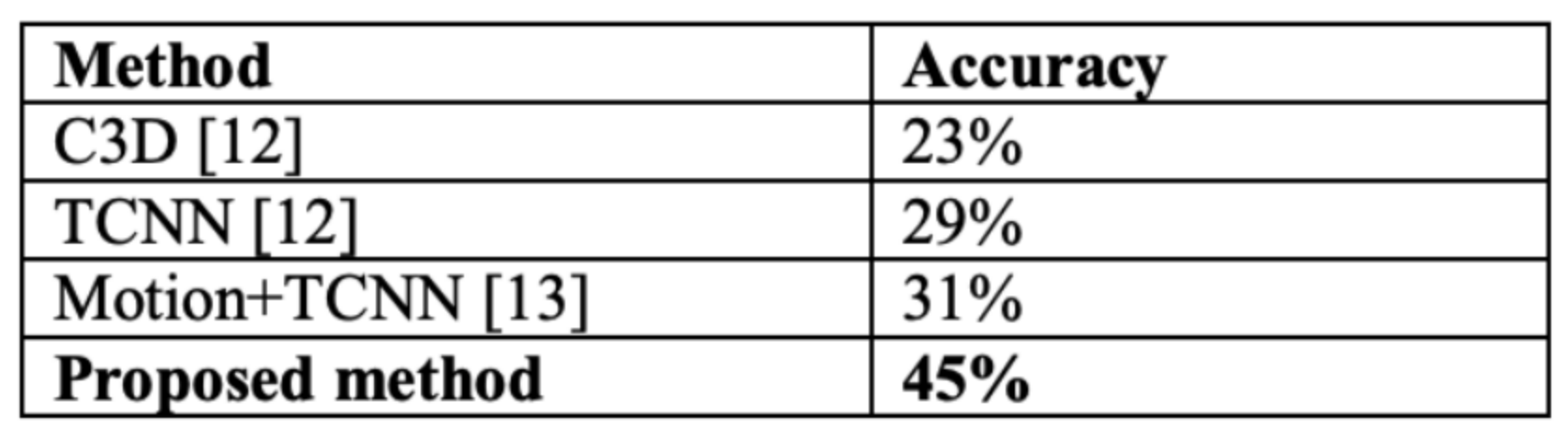}
    \caption{Supervised Model Multiclass Accuracy Comparison of UCF Crime dataset}
    \label{fig:multiclassComparison}
\end{figure}

\begin{figure}[ht]
    \centering
    \includegraphics[width=0.96\columnwidth]{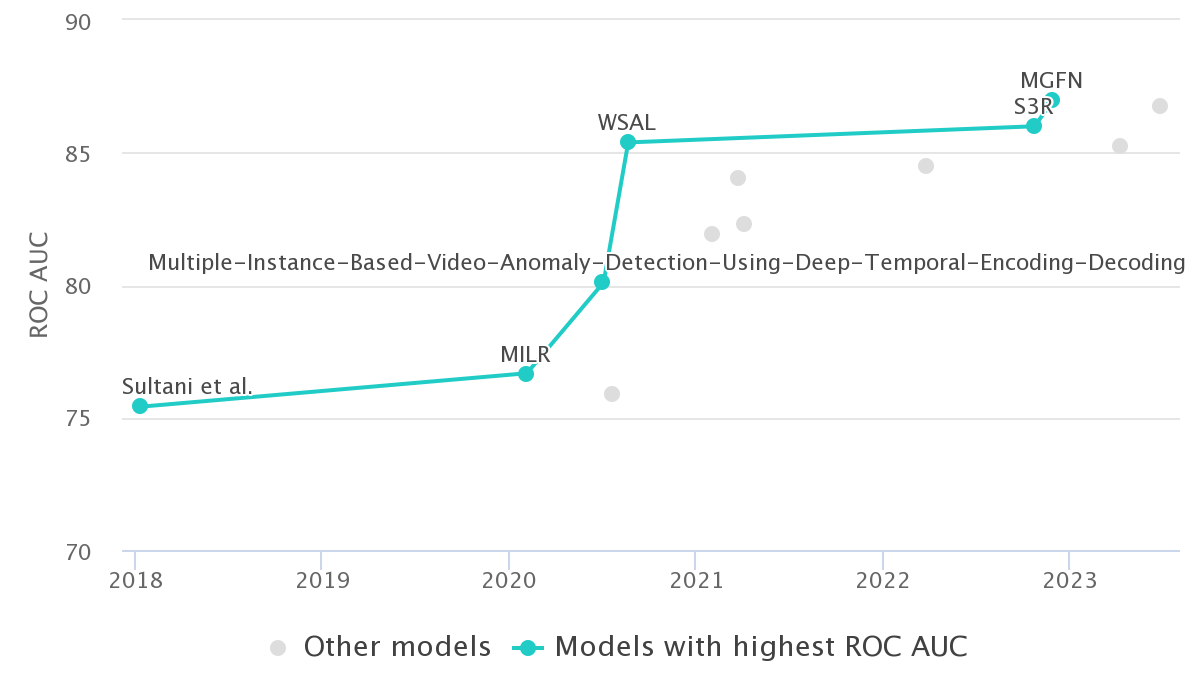}
    \caption{Supervised Model Anomaly Detection UCF Crime dataset}
    \label{fig:rocComparison}
\end{figure}

As shown in figures \ref{fig:multiclassComparison} and  \ref{fig:rocComparison}, initial testing of Ethosight zero-shot performance on the UCF Crime database does not achieve state-of-the-art performance when compared to supervised models. However, even at the current accuracy levels, Ethosight's ability to continuously and cumulatively learn and produce increasingly more accurate models may help narrow this performance gap. It should also be noted that, due to time constraints, these baseline numbers do not yet include the full performance expected from Ethosight's full capabilities. 

\begin{figure}[ht]
\centering
\includegraphics[width=0.45\textwidth]{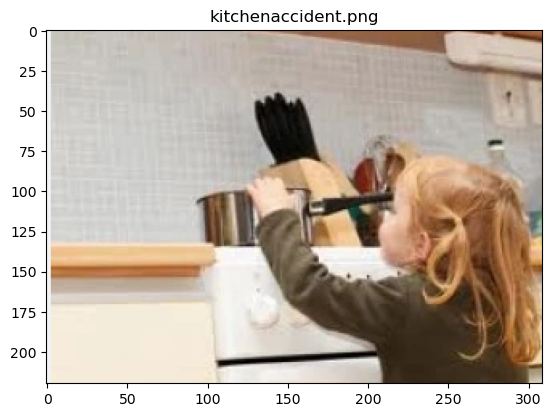}
\caption{Dangerous kitchen event detected by Ethosight.}
\label{fig:kitchenaccident_event}
\end{figure}

Due to the lack of datasets for detecting nuanced health, safety, and security related events and behaviors from single images, we are also releasing a new dataset containing 20 classes of interest including child in danger [figure: \ref{fig:kitchenaccident_event}].

\subsubsection{Semantic Label Expansion and its Impact}

The introduction of semantic label expansion in Ethosight plays a pivotal role in improving the signal-to-noise ratio, thereby enhancing classification performance. By expanding initial ground truth labels with positive, negative, and differentiating evidence, Ethosight is equipped with a richer context to make informed decisions.

\begin{figure}[ht]
    \centering
    \includegraphics[width=0.7\columnwidth]{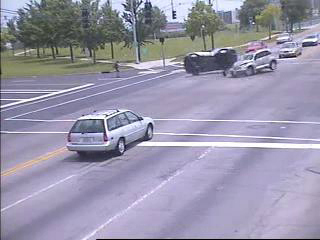}
    \caption{Snapshot from the UCF Crime dataset depicting a road accident.}
    \label{fig:roadAccident}
\end{figure}

\begin{figure}[ht]
    \centering
    \includegraphics[width=0.9\columnwidth]{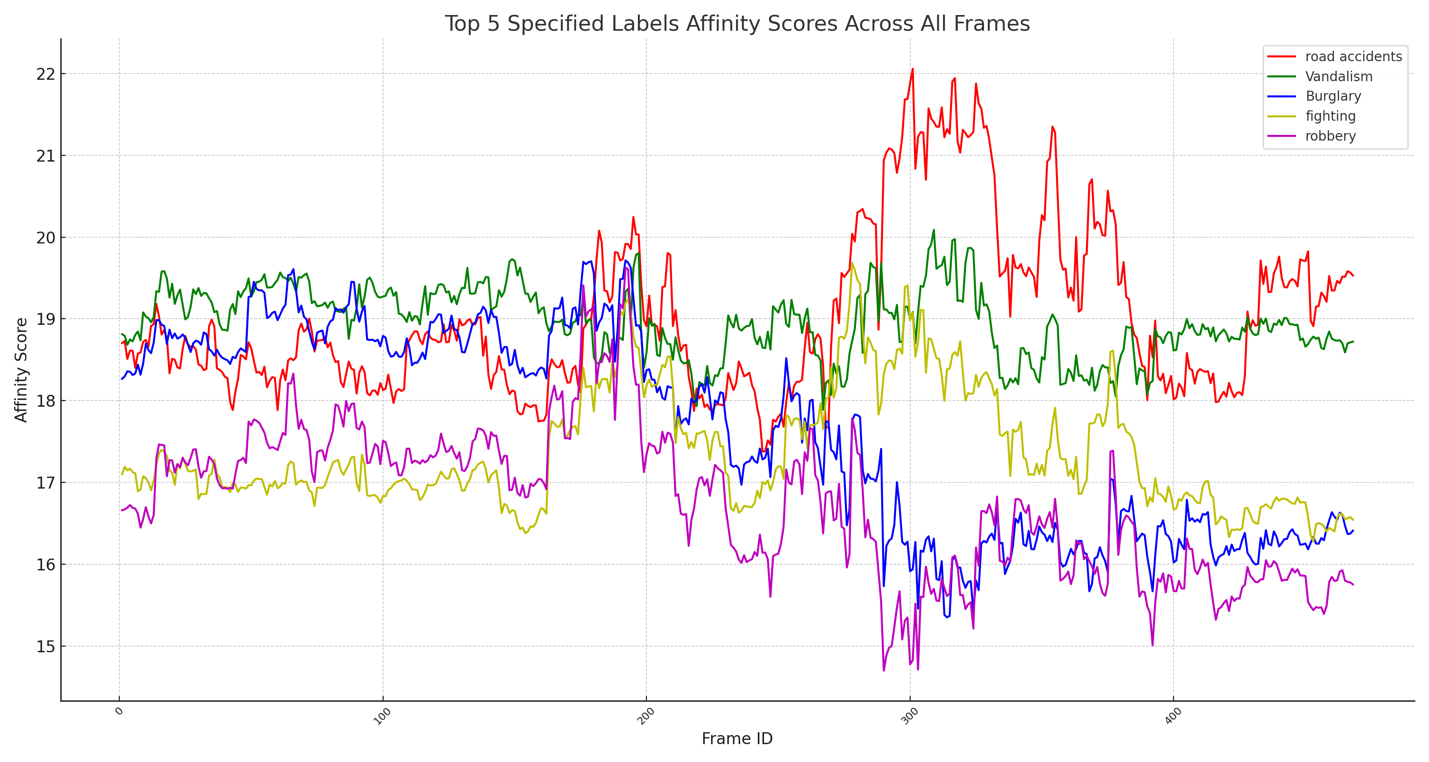}
    \caption{Ethosight without semantic label optimization}
    \label{fig:roc_groundtruth}
\end{figure}

\begin{figure}[ht]
    \centering
    \includegraphics[width=0.9\columnwidth]{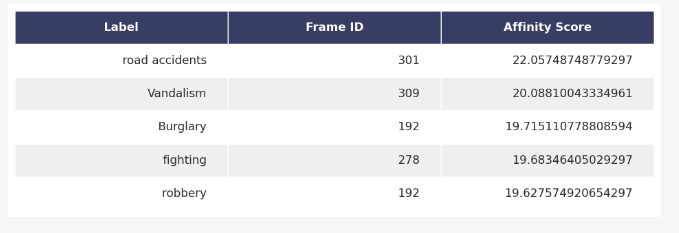}
    \caption{Ethosight without semantic label optimization}
    \label{fig:groundTruthMetrics}
\end{figure}

\begin{figure}[ht]
    \centering
    \includegraphics[width=0.9\columnwidth]{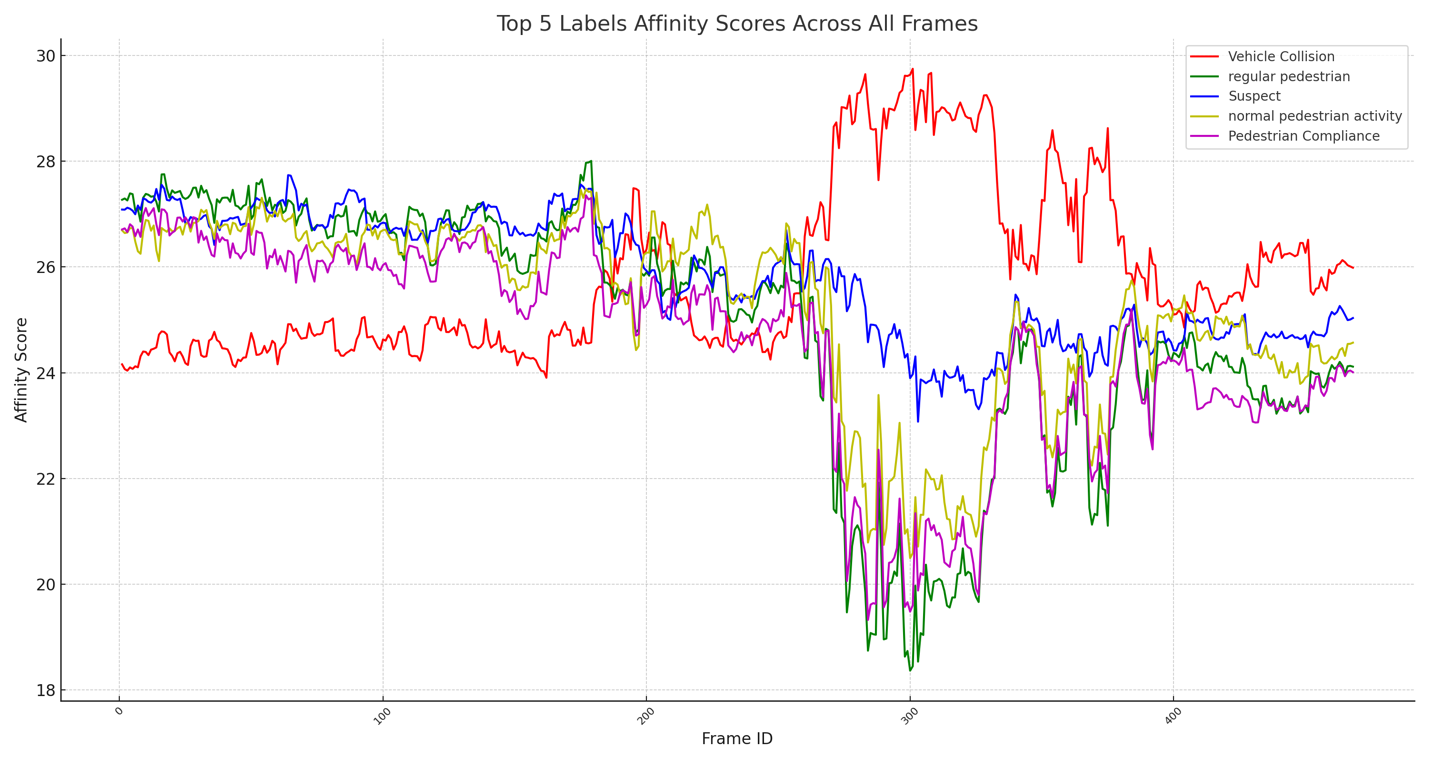}
    \caption{Ethosight with semantic label optimization.}
    \label{fig:roc_semantic}
\end{figure}

\begin{figure}[ht]
    \centering
    \includegraphics[width=0.9\columnwidth]{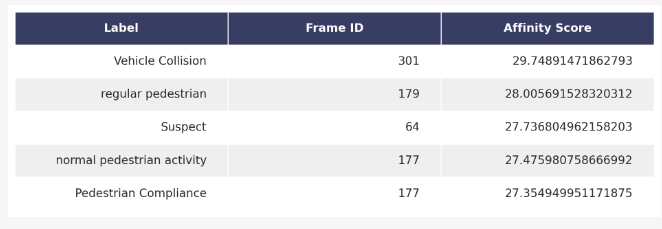}
    \caption{Ethosight with semantic label optimization}
    \label{fig:semanticLabelExpansionMetrics}
\end{figure}

Comparative evaluation using figures \ref{fig:roc_groundtruth} and \ref{fig:roc_semantic} indicate a significant performance boost post semantic label expansion. It underscores the potential of Ethosight's continuous cumulative learning, to match the accuracy of traditional supervised models.

\begin{table}[ht]
\centering
\begin{tabular}{|c|c|}
\hline
\textbf{Label} & \textbf{Affinity Score} \\ 
\hline
child in danger & 0.4346351623535156 \\ 
\hline
cooking & 0.26900413632392883 \\ 
\hline
burglary & 0.0736141949892044 \\ 
\hline
fire\_prevention & 0.07326945662498474 \\ 
\hline
gas\_leak\_detection & 0.06314415484666824 \\ 
\hline
repairing & 0.017247511073946953 \\ 
\hline
vomiting & 0.008393174037337303 \\ 
\hline
fire\_safety & 0.0065605794079601765 \\ 
\hline
decorating & 0.006369316019117832 \\ 
\hline
buying & 0.0054978723637759686 \\ 
\hline
peeping tom & 0.004484186880290508 \\ 
\hline
fire\_investigation & 0.0030823920387774706 \\ 
\hline
child abuse & 0.0020500756800174713 \\ 
\hline
oven & 0.0017948184395208955 \\ 
\hline
monitor & 0.0016760976286605 \\ 
\hline
\end{tabular}
\caption{Kitchen Event Labels and Affinity Scores}
\label{table:2}
\end{table}

\subsection{Ethosight Open Source}
Ethosight is an open source project and all code, data, and other digital artifacts are available as open source under the Apache 2.0 license. More details can be found at the following URL: \url{https://research.cisco.com/research-projects/deep-vision}.

%% file: sections/conclusionAndNextSteps.tex
\section{Conclusion and Future Work}

\subsection{Conclusion}
Ethosight, as part of the Deep Vision framework, introduces a novel approach to enabling real-time video analytics through a zero-shot learning paradigm. Uniquely, Ethosight begins its analysis from a blank slate, relying solely on natural language input or predefined keyword labels, without the requirement for any training data. The system integrates various reasoning engines, which can include logic-based models like OpenNARS, or other symbolic reasoning systems, to enable adaptive and continuous learning. Ethosight effectively employs joint embeddings to process image data, with ImageBind emerging as the most promising implementation to date.

It is important to acknowledge that Ethosight's accuracy, as a zero-shot learning system, does not currently reach the state-of-the-art performance of supervised models on tasks such as multi-class classification in the UCF Crime dataset. Unlike supervised models, Ethosight operates without the need for any supervised training data. Its unique strength lies in its ability to learn symbolically, allowing for continuous, cumulative learning that is adaptable and can be extended over time without requiring extensive retraining. This zero-shot nature and symbolic reasoning are fundamental features that distinguish Ethosight and underline its promise as an evolving solution for real-time video analytics.

\subsection{Future Work}
Potential directions for future work include:
\begin{itemize}
    \item \textbf{Continuous Cumulative Learning:} A central focus of Ethosight is on continuous and cumulative learning, wherein the model evolves and refines its knowledge over time without requiring complete retraining. This learning is not tied to a static dataset but progresses as new data and contexts are encountered. Ethosight aims to build a growing, machine-generated ontology that becomes increasingly sophisticated as it interacts with more data, enabling the development of more accurate and portable models between different systems and applications.
    \item \textbf{Spatio-Temporal Focus of Attention:} Developing capabilities for Ethosight to dynamically direct its focus towards specific spatio-temporal regions within the input data. This focus of attention aims to enable the creation of more detailed and contextually relevant machine-generated symbolic knowledge. It will allow Ethosight to identify and analyze critical regions of a scene more effectively, leading to more insightful and precise analytics.
    \item \textbf{Symbolic Reasoning and Label Generation:} Further integration with symbolic reasoning engines, such as OpenNARS, to enhance Ethosight’s capabilities in inductive and abductive reasoning. This integration aims to allow Ethosight to explore reasoning around inheritance relations for label generation, thereby optimizing label spaces in a more cost-efficient manner.
    \item \textbf{Self-supervised Behavior Learning:} Experimenting with self-supervised approaches that may enable Ethosight to automatically identify and adapt to significant patterns and anomalies in the input data, based on contextually relevant criteria. This could allow the model to further refine its predictions and adapt to new, unseen data more effectively.
    \item \textbf{Multi-Object Tracking Integration:} Planning to incorporate advanced tracking technologies, such as hybrid AI multi-object tracking, into Ethosight's framework. This integration is aimed at enhancing Ethosight's real-time analytics capabilities, allowing it to more effectively track and analyze multiple objects within a scene.
\end{itemize}